\documentclass[conference]{./support/ieeeconf}
\pdfoutput=1 
\usepackage{times}
\usepackage{amsmath}
\usepackage[pdftex]{graphicx}
\usepackage{caption}
\usepackage{subcaption}
\usepackage{amsmath,amssymb,amsopn,amstext,amsfonts}
\usepackage{cancel}
\usepackage[space]{cite}
\usepackage{pdfsync}
\usepackage{color}
\usepackage{algorithm}
\usepackage{algorithmic}
\usepackage{url}
\usepackage{booktabs}
\usepackage{threeparttable}
\usepackage[linkcolor=black,citecolor=black,urlcolor=black,colorlinks=true]{hyperref}
\usepackage{multirow}
\usepackage{makecell}
\usepackage[outdir=./epstopdf/]{epstopdf}
\usepackage{graphicx}
\usepackage{array}
\usepackage{verbatim}
\usepackage{makecell}
\usepackage{booktabs}
\usepackage{siunitx}
\usepackage{multirow}
\usepackage{afterpage}
\usepackage{tabularx}
\usepackage{rotating}
\usepackage{xcolor}
\usepackage{balance}
\usepackage{xcolor}
\usepackage{hyperref}
\usepackage{color}
\usepackage{hyperref}
\hypersetup{
    urlcolor=blue,
}
\colorlet{urlcolor}{red}

\IEEEoverridecommandlockouts

\graphicspath{{./figure/}}

\DeclareGraphicsExtensions{.pdf,.png,.jpg,.eps,.PNG}
\title{\LARGE \bf ParkingE2E: Camera-based End-to-end Parking Network, from Images to Planning}
\author{Changze Li, Ziheng Ji, Zhe Chen, Tong Qin$^*$, and Ming Yang
		\thanks{
					All authors are with the Global Institute of Future Technology, Shanghai Jiao Tong University, Shanghai, China.
		{\tt\small  \{changze, jiziheng, Zhe\_Chen, qintong, mingyang\}@sjtu.edu.cn}. 
		{  $^*$ is the corresponding author}.
	}}

\begin{document}

\maketitle
\thispagestyle{empty}
\pagestyle{empty}

\begin{abstract}
Autonomous parking is a crucial task in the intelligent driving field.
Traditional parking algorithms are usually implemented using rule-based schemes.
However, these methods are less effective in complex parking scenarios due to the intricate design of the algorithms.
In contrast, neural-network-based methods tend to be more intuitive and versatile than the rule-based methods.
By collecting a large number of expert parking trajectory data and emulating human strategy via learning-based methods, the parking task can be effectively addressed.
In this paper, we employ imitation learning to perform end-to-end planning from RGB images to path planning by imitating human driving trajectories.
The proposed end-to-end approach utilizes a target query encoder to fuse images and target features, and a transformer-based decoder to autoregressively predict future waypoints.
We conduct extensive experiments in real-world scenarios, and the results demonstrate that the proposed method achieved an average parking success rate of 87.8\% across four different real-world garages.
Real-vehicle experiments further validate the feasibility and effectiveness of the method proposed in this paper. 
The code can be found at: \url{https://github.com/qintonguav/ParkingE2E}.

\end{abstract}

\section{Introduction}
Intelligent driving involves three main tasks: urban driving, highway driving, and parking maneuvers. 
Automated valet parking (AVP) and auto parking assist (APA) systems, crucial parking tasks within intelligent driving, offer significant improvements in parking safety and convenience.
However, mainstream parking methods \cite{qin2020avp} are often rule-based, requiring the entire parking process to be decomposed into multiple stages such as environmental perception, mapping, slot detection, localization and path planning.
Due to the intricate nature of these complex model architectures, they are more susceptible to encountering difficulties in tight parking spots or intricate scenarios.

End-to-end (E2E) autonomous driving algorithms \cite{chitta2022transfuser, chen2020learning, codevilla2019exploring, wu2022trajectory, hu2023planning} mitigate cumulative errors across modules by integrating perception, prediction, and planning components into a unified neural network for joint optimization.
The application of end-to-end algorithms to parking scenarios helps decrease the dependence of parking systems on manually designed features and rules, providing a comprehensive, holistic, and user-friendly solution.

While end-to-end autonomous driving has demonstrated significant advantages, most of the research has concentrated on simulation \cite{dosovitskiy2017carla} without validating the algorithm's real-world effectiveness.
In contrast to the complexity of urban environments and the hazards of highway driving, parking scenarios are characterized by low speeds, confined spaces, and high controllability. These features provide a feasible pathway for incrementally deploying end-to-end autonomous driving capabilities in vehicles. We develop an end-to-end parking neural network and validate the algorithm's feasibility in real-world parking situations.

\begin{figure}[t]
	\centering
	\includegraphics[width=0.5\textwidth]{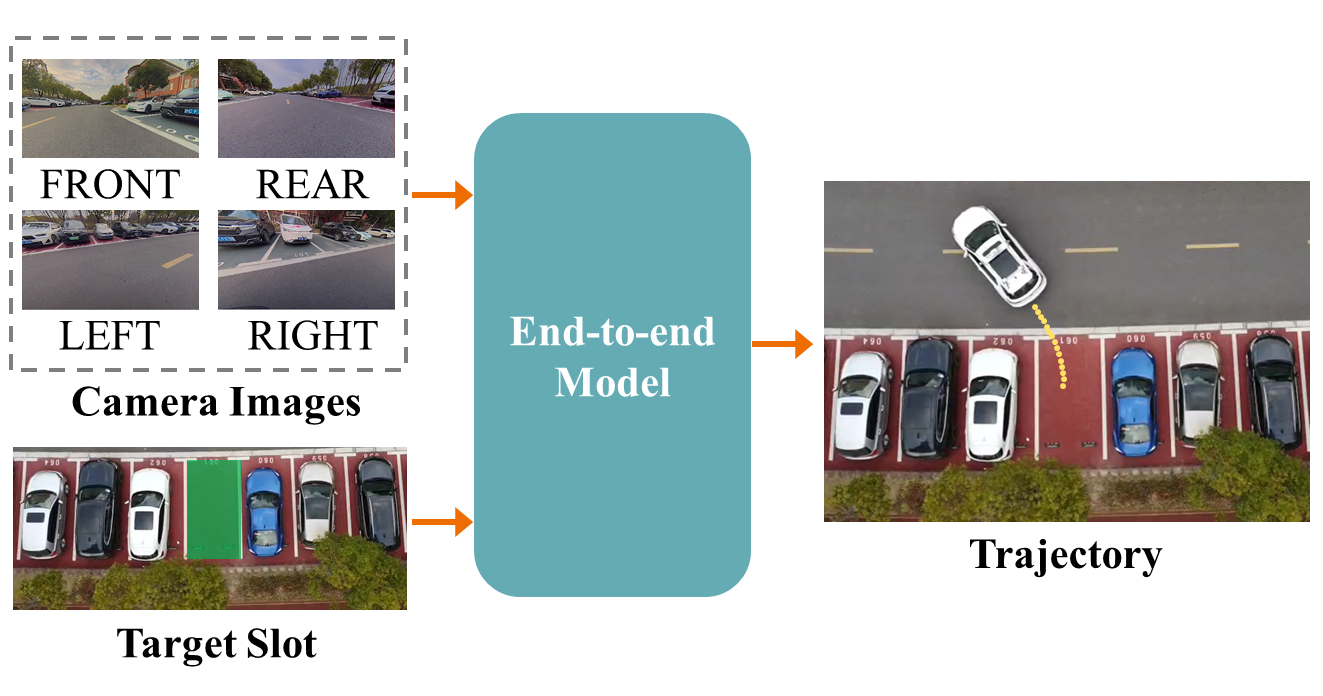}
	\caption{Illustration of the overall workflow. Our model takes the surround-view camera images and the target slot as inputs and outputs the predicted trajectory waypoints, which are later executed by the controller. Supplementary video material is available at: \url{https://youtu.be/urOEHJH1TBQ}.}
    \label{structure_overview}
    \vspace{-1.0cm}
\end{figure}

This work extends our previous work E2E-Carla \cite{E2EAPA} by presenting an imitation-learning-based end-to-end parking algorithm, which has been successfully deployed and evaluated in real environments.
The algorithm takes in surround-view images captured by on-board cameras, predicts future trajectory results, and executes control based on the predicted waypoints.
Once the user designates a parking slot, the end-to-end parking network collaborates with the controller to automatically maneuver the vehicle into the parking slot until it is fully parked.
The contributions of this paper are summarized as follows:

\begin{itemize}
    \item {
    We designed an end-to-end network to perform parking task.
    The network converts the surround view images into Bird's Eye View (BEV) representation, and fuses it with the target parking slot features by employing the target features to query the image features.
    Due to the sequential nature of the trajectory points, we utilize an autoregressive approach based on transformer decoder to generate trajectory points.
    }
    \item {
    We deployed the end-to-end model on \textbf{real vehicles} for testing and verified the feasibility and generalizability of the network model for parking across various \textbf{real-world scenarios}, offering an effective solution for end-to-end network deployment.
    }
\end{itemize}

\section{literature review}
\subsection{BEV Perception}
BEV representation offers at least two advantages over perspective representation. First, it easily integrates inputs from different modalities due to its clear physical interpretability.
Secondly, the BEV view avoids perspective distortion issues, thereby reducing the complexity of downstream tasks such as planning.
In recent years, the BEV representation has seen widespread adoption in perception systems.
Unlike previous deep learning-based perception algorithms comprising a feature extraction module and a task head module, BEV perception incorporates an additional viewpoint conversion module alongside these two modules. This conversion module facilitates the transformation between the sensor view and the Bird's Eye View (BEV).

LSS \cite{LSS} utilizes BEV perception for detection and segmentation. This method acquires BEV features by estimating the depth distribution at each pixel of the feature map and projecting it onto the BEV plane.
DETR3D \cite{wang2022detr3d} follows the basic paradigm of DETR \cite{carion2020end} and employs sparse queries for 3D object detection.
PETR \cite{liu2022petr} adds 3D positional embedding, which provides 3D positional information to 2D features, aiming to the neural network to implicitly learn depth.
BEVFormer \cite{li2022bevformer} adopts BEV queries for perception, and incorporates spatial cross-attention and temporal self-attention mechanisms to boost performance.
BEVDepth \cite{li2023bevdepth} builds upon LSS and utilizes LiDAR points for depth supervision during  training to enhance the depth estimation quality, thereby improving BEV perception performance.
BEVFusion \cite{liang2022bevfusion} extracts BEV features from both cameras and LiDAR data and fuses them in the BEV space.

\subsection{End-to-end Autonomous Driving}
In contrast to the traditional module-based autonomous driving solutions, the end-to-end paradigm \cite{endtoend_summary, chib2023recent} can mitigate accumulated errors, prevent information loss across module and minimize redundant computations. 
Consequently, it has emerged as a popular and prominent research topic in the field of autonomous driving tasks.

The research on end-to-end driving initially focused on autonomous urban driving tasks.
Chauffeur{Net} \cite{chauffeurnet}, an imitation-learning-based end-to-end method, learned effective driving strategies from expert data.
Many methods have adopted an encoder-decoder framework that extracted BEV features from sensors and then utilizes GRU (Gate Recurrent Unit) decoder to predict waypoints in an autoregressive manner, such as Transfuser \cite{chitta2022transfuser, transfuser}, Interfuser \cite{interfuser}, and NEAT \cite{NEAT}.
Besides, CIL \cite{CIL} and CILRS \cite{CILRS} developed a neural network that directly maps front-view images, current measurements, and navigation commands into control signals without a separate PID controller.
MP3 \cite{casas2021mp3} and UniAD \cite{hu2023planning} propose a modular design but jointly optimize all components in an end-to-end manner.

In recent years, end-to-end networks have been developed for parking scenarios.
Rathour et al. \cite{rathour2018vision} proposed a two-stage learning framework to predict the steering angles and gears from images.
In the first stage, the network predicts an initial estimate of a sequence of steering angles.
In the second stage, an LSTM (Long Short-Term Memory) network are used to estimate the optimal steering angles and gears.
Li et al. \cite{li2018end} trained a CNN (Convolutional Neural Network) on rear-view images to automatically control steering angle and velocity.
ParkPredict \cite{ParkPredict} proposed a parking slot and waypoints prediction network based on a CNN-LSTM architecture.
In the following work, ParkPredict+ \cite{ParkPredict+} designed a transformer and CNN-based model to predict future vehicle trajectories based on intent, image, and historical trajectory.

Existing end-to-end autonomous driving methods often demand substantial computational resources, pose training challenges, and face difficulties in real-vehicle deployment.
On the other hand, parking approaches exemplified by ParkPredict primarily focus on prediction from aerial imagery, which differs from our task.
Our method propose an end-to-end parking planning network that utilizes an autoregressive transformer decoder to predict future waypoints from BEV features extracted from RGB images and target slot.

\begin{figure*}[t]
	\centering
\includegraphics[width=1\textwidth]{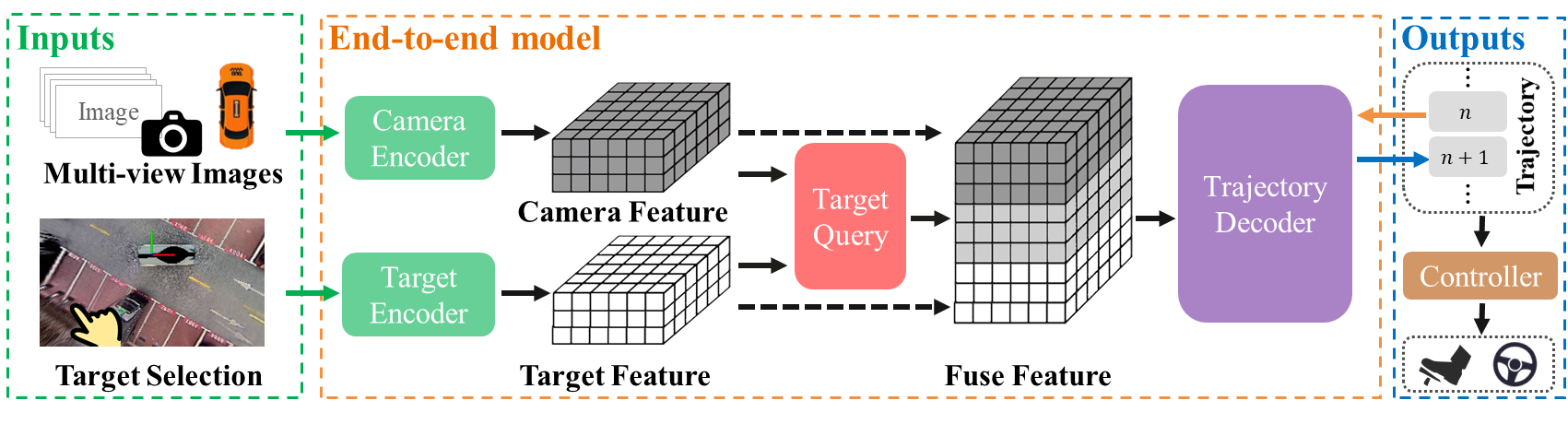}
	\caption{Overview of our method. Multi-view RGB images are processed and the image features are transformed into BEV representation. The target slot is used to generate the BEV target features. We fuse target features and image BEV features using target query. Then we obtain the predicted trajectory points one by one using the autoregressive transformer decoder.}
	\label{neural_parking_overview}
\vspace{-0.3cm}
\end{figure*}

\section{methodology}
\subsection{Preliminaries: Problem Definition}
We use end-to-end neural network $\mathcal{N_\theta}$ to imitate expert trajectories for training, defining the dataset:
\begin{equation}
    \mathcal{D}=\{(I_{i,j}^k, P_{i,j}, S_i)\},
\end{equation}
where trajectory index $i\in[1,M]$, trajectory points index $j\in[1,N_i]$, camera index $k\in[1, R]$, RGB image $I$, trajectory point $P$ and target slot $S$.
Reorganize the dataset into:
\begin{equation}
    \mathcal{T}_{i,j}=\{P_{i,\min(j+b, N_i)}\}_{b = 1,2,\dots,Q},
\end{equation}
and
\begin{equation}
    \mathcal{D'}=\{(I_{i,j}^k, \mathcal{T}_{i,j}, S_i)\},
\end{equation}
where $Q$ denotes the length of the predicted trajectory points and $R$ denotes the number of RGB cameras.

The optimization goals for the end-to-end network are as follows:
\begin{equation}
\theta' = \mathop{\arg\min}\limits_{\theta}\mathbb{E}_{(I, \mathcal{T}, S)\sim \mathcal{D'}}[\mathcal{L}(\mathcal{T}, \mathcal{N_\theta}(I, S)) ],
\end{equation}
where $\mathcal{L}$ denotes the loss function.

\subsection{Camera-based End-to-end Neural Planner}
\subsubsection{\textbf{Overview}}
As shown in Fig. \ref{neural_parking_overview}, we developed an end-to-end neural planner that takes RGB images and a target slot as input.
The proposed neural network includes two main parts: an input encoder and an autoregressive trajectory decoder.
With the input of RGB images and the target slot, the RGB images are transformed to BEV features.
Then, the neural network fuses BEV features with the target slot and generates the next trajectory point in an autoregressive way using transformer decoder.
\subsubsection{\textbf{Encoder}}
We encode the inputs in the BEV view.
The BEV representation provides a top-down view of the vehicle's surrounding environment, allowing the ego-vehicle to detect parking slots, obstacles, and markings.
At the same time, the BEV view provides a consistent viewpoint representation across various driving perspectives, thereby simplifying the complexity of trajectory prediction.

\textbf{Camera Encoder}
At the beginning of the BEV generation pipeline, we first utilize EfficientNet \cite{efficientnet} to extract image features $\mathcal{F}_{img} \in \mathbb{R}^{{C}\times{H_{img}}\times{W_{img}}}$ from RGB inputs. 
Inspired by LSS \cite{LSS}, We learn a depth distribution $d_{dep} \in \mathbb{R}^{{D}\times{H_{img}}\times{W_{img}}}$ of image features and lift each pixel into 3D space.
We then multiply the predicted depth distribution $d_{dep}$ and image feature $\mathcal{F}_{img}$ to obtain the image feature with depth information.
With the camera extrinsics and intrinsics, image features are projected into BEV voxel grid to generate camera features  $\mathcal{F}_{cam} \in \mathbb{R}^{{C}\times{H_{cam}}\times{W_{cam}}}$. 
The range of BEV features in the $x$-direction is denoted as $[-R_x , R_x ]\mathrm{m}$, where $\mathrm{m}$ denotes meters, and the range in the $y$-direction is denoted as $[-R_y , R_y]\mathrm{m}$.


\textbf{Target Encoder}
In order to align the target slot with the camera feature $\mathcal{F}_{cam}$, we generate a target heat map in the BEV space as the input of target encoder based on the specified parking slot location. Subsequently, we extract the target slot features $\mathcal{F}_{target}$ using a deep CNN neural network to obtain the same dimension with $\mathcal{F}_{cam}$.
During the training, the target parking slot is determined by the end points of the human driving trajectory.
\begin{figure}[t]
    \includegraphics[width=0.5\textwidth]{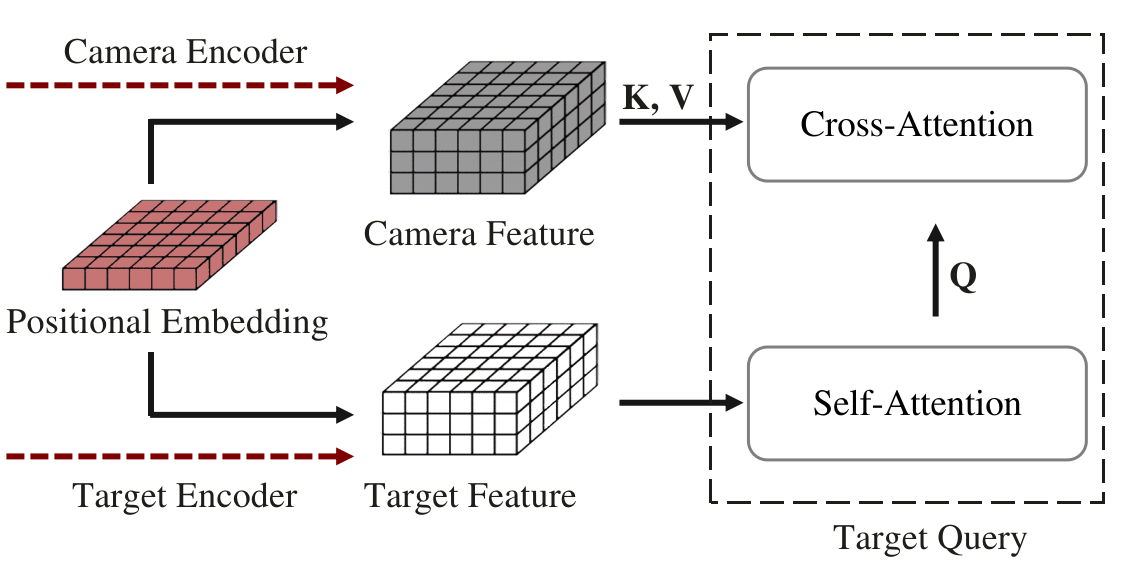}
	\caption{The architecture of the target query illustrates that we add the same positional encoding to the target feature and camera feature to establish the spatial relationship between the two types of features.}
	\label{target_query}
     \vspace{-0.5cm}
\end{figure}

\textbf{Target Query}
By aligning the camera features $\mathcal{F}_{cam}$ and the target encoding features $\mathcal{F}_{target}$ in BEV space and using the target feature to query the camera feature via the cross-attention mechanism, we can effectively fuse the two modalities.
The positional encoding ensures the spatial correspondence is maintained between camera features and target features when associating the features at the specific BEV location. 
Utilizing $\mathcal{F}_{target}$ as the query, camera feature $\mathcal{F}_{cam}$ as the key and the value and employing the attention mechanism, we obtain the fused feature $\mathcal{F}_{fuse}$.



\subsubsection {\textbf{Decoder}}
Many end-to-end planning studies \cite{transfuser,interfuser,NEAT} have employed a GRU decoder to predict next points from a high-dimensional feature vectors in an autoregressive way.
However, the high-dimensional vectors of features lack a global receptive field.
Taking inspiration from Pix2seq\cite{Pix2seq}, we approach trajectory planning as a sequence prediction problem using a transformer decoder. This involves autoregressive, step-by-step prediction of the trajectory points.
Our approach effectively combines low-dimensional trajectory points with high-dimensional image features.

\textbf{Trajectory Serialization}
Trajectory serialization represents trajectory points as discrete tokens.
By serializing the trajectory points, the position regression can be converted into token prediction.
Subsequently, we can leverage a transformer decoder to predict the trajectory point $(P_{ij}^x, P_{ij}^y)$ in the ego vehicle's coordinate system, we utilize the following serialization method:
\begin{equation}
\mathrm{Ser}(P_{i,j}^x) = \lfloor \frac{P_{i,j}^x + R_x}{2 R_x} \rfloor \times N_t,
\end{equation}
and
\begin{equation}
\mathrm{Ser}(P_{i,j}^y) = \lfloor \frac{P_{i,j}^y + R_y}{2 R_y} \rfloor \times N_t,
\end{equation}
where $N_t$ represents the maximum value that can be encoded by a token in the sequence and the symbol for serializing trajectory points is denoted as $\mathrm{Ser}(\cdot)$.
$R_x$ and $R_y$ represent the maximum values of the predicted range in the $x$ and $y$ directions, respectively.


After serialization, the $i$-th trajectory can be expressed as follow:

\begin{equation}
\begin{split}
[\mathrm{BOS}, \mathrm{Ser}(P_{i,1}^x), \mathrm{Ser}(P_{i,1}^y), ..., \\
\mathrm{Ser}(P_{i,N_i}^x), \mathrm{Ser}(P_{i,N_i}^y), \mathrm{EOS}],
\end{split}
\end{equation}
where $\mathrm{BOS}$ represents the start flag and $\mathrm{EOS}$ represents the end flag.

\textbf{Trajectory Decoder}
The BEV features serve as the key and the value, while the serialization sequence is utilized as the query to generate trajectory points using a transformer decoder in an autoregressive manner.
During training, we add positional embedding to the sequence points and implement parallelization by masking unknown information.
During inference process, given the $\mathrm{BOS}$ token, then the transformer decoder predicts following points in sequence. Then we append the predicted point to the sequence for the next step repeating this process until encountering $\mathrm{EOS}$ or reaching the specified number of predicted points.
\subsection{Lateral and Longitudinal Control}
During the control process, the parking start moment denoted as $t_0$, is used as the starting time to predict the path  $\mathcal{T}_{t_0}=\mathcal{N_{\theta'}}(I_{t_0},S)$ based on the end-to-end neural planner and the relative pose from the initial moment $t_0$ to the current moment $t$ can be obtained by the localization system, denoted as $ego_{t_0 \to t}$.
The target steering angle $\mathcal{A}^{tar}$ can be obtained using the RWF (Rear-wheel Feedback) method, which can be expressed as follows:
\begin{equation}
\mathcal{A}_t^{tar} = \mathbf{\mathrm{RWF}}(\mathcal{T}_{t_0}, ego_{t_0 \to t}).
\end{equation}
According to the speed feedback $\mathcal{V}^{feed}$ and steer feedback $\mathcal{A}^{feed}$ from the chassis, as well as the target speed $\mathcal{V}^{tar}$ from the setting and the target steer $\mathcal{A}^{tar}$ from the calculation, cascade PID controller is utilized to achieve lateral and longitudinal control.
After a new predicted trajectory is generated, $\mathcal{T}_{t_0}$ and $ego_{t_0 \to t}$ are reset, eliminating the necessity of relying on global localization throughout the entire vehicle control process.
\section{Experiments}
\label{sec:Experiments}
\subsection{Dataset Collection}
\begin{figure}[h]
	\centering
    \includegraphics[width=0.48\textwidth]{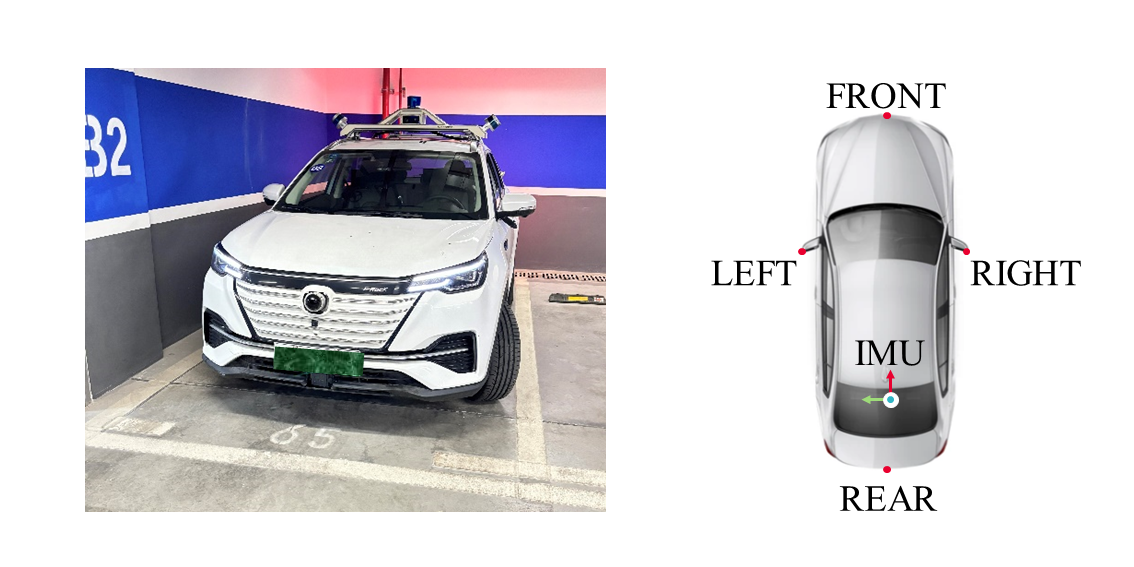}
	\caption{We use a Changan vehicle as the experimental platform. The vehicle utilizes Intel NUC devices to execute model inference and control.}
	\label{platform}\
\vspace{-0.5cm}
\end{figure}

\begin{figure}[htbp]
	\centering
	\includegraphics[width=0.47\textwidth]{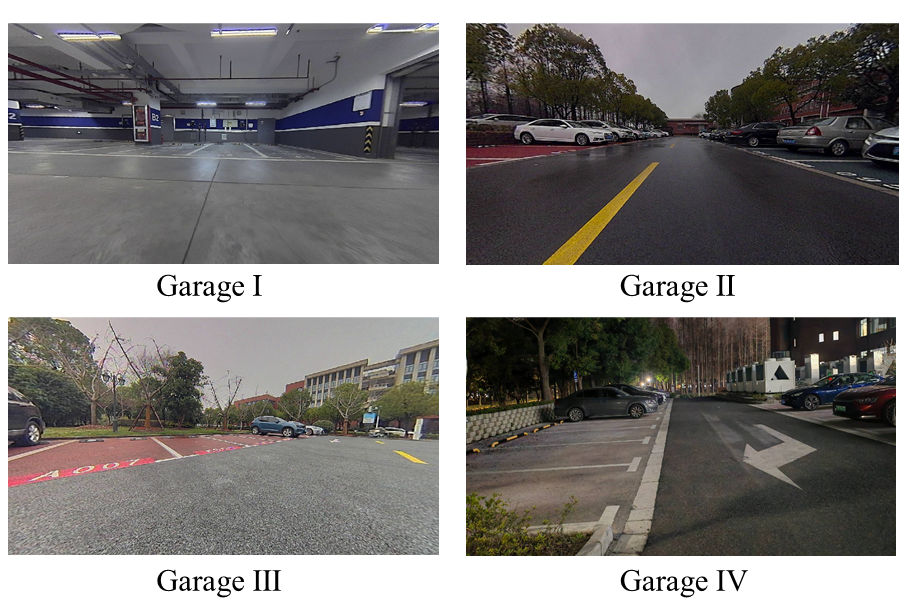}
	\caption{Several different garages are utilized for training and testing the system. Some of the parking slot data from Garage \uppercase\expandafter{\romannumeral 1} and \uppercase\expandafter{\romannumeral 2} are used for training. While the 
 remaining parking slot data from Garage \uppercase\expandafter{\romannumeral 1} and \uppercase\expandafter{\romannumeral 2} that are not involved in training as well as all collected slot data from Garage \uppercase\expandafter{\romannumeral 3} and \uppercase\expandafter{\romannumeral 4} are used for testing.}
	\label{scene}
\vspace{-0.2cm}
\end{figure}

The datasets are collected using vehicle-mounted devices.
To facilitate comprehensive visual perception and trajectories, surround-view cameras are employed to capture RGB images.
Concurrently, dead reckoning techniques are integrated, leveraging sensor data fusion algorithms to achieve robust and accurate vehicle localization. 
The layout of the experimental platform and the sensors used are shown in Fig. \ref{platform}.
The data is collected across various scenarios for parking, including underground and ground garages, as shown in Fig. \ref{scene}.
Collecting data from diverse environments helps enhance the generalization capability of neural networks.

\setlength{\extrarowheight}{1.5pt} 
\begin{table*}[t]
\centering
\caption{Quantitative results of parking performance tests in different parking scenarios}

\label{tab:vehicle_close_test}
\begin{tabular}{c|c|cccccccc}
\toprule
Garage & Scene & PSR (\%) $\uparrow$ & NSR (\%) $\downarrow$ & PVR (\%) $\downarrow$ & APE ($\mathrm{m}$) $\downarrow$ & AOE ($\mathrm{deg}$) $\downarrow$ & APT ($\mathrm{s}$) $\downarrow$ & APS $\uparrow$\\
\midrule
         & Scene A & 70.3 & 3.7 & 62.9 & 0.59 & 7.2 & 64 & 51.5 \\
Garage \uppercase\expandafter{\romannumeral 1} &  Scene B & 90.7 & 0.0 & 38.8 & 0.58 & 3.9 & 70 & 81.5 \\
         & Scene C & 83.3 & 8.3 & 58.3 & 0.62 & 5.8 & 60 & 63.5  \\
\midrule
         & Scene A & 83.3 & 0.0 & 50.0 & 0.35 & 10.0 & 66 & 58.0 \\
Garage  \uppercase\expandafter{\romannumeral 2} &  Scene B & 91.6 & 0.0 & 58.3 & 0.47 & 6.5 & 63 & 69.7 \\
         & Scene C & 81.2 & 0.0 & 50.0 & 0.60 & 6.8 & 64 & 64.6 \\
\midrule
         & Scene A & 95.8 & 0.0 & 33.3 & 0.20 & 2.5 & 51 & 88.6 \\
Garage  \uppercase\expandafter{\romannumeral 3} &  Scene B & 100.0 & 0.0 & 25.0 & 0.34 & 5.2 & 50 & 83.1 \\
         & Scene C & 91.6 & 8.3 & 41.6 & 0.93 & 7.4 & 55 & 68.6 \\
\midrule
         & Scene A & 94.3 & 0.0 & 16.7 & 0.50 & 3.0 & 81 & 82.1 \\
Garage  \uppercase\expandafter{\romannumeral 4} &  Scene B & 88.7 & 0.0 & 11.1 & 1.14 & 7.4 & 86 & 75.1 \\
         & Scene C & 83.3 & 16.6 & 33.2 & 0.56 & 3.5 & 96 & 65.9 \\

\bottomrule
\end{tabular}
\end{table*}

\subsection{Implement Details}
During the training process, surround-view camera images (the number of cameras $R$ is 4) are used as input, and the target parking space is determined by some points at the end of the parking.
The trajectory sequence points are used to supervise the end-to-end prediction results.

In the inference process, the target parking slot is selected by using ``2D-Nav-Goal" in the RViz interface software to obtain the target parking slot.
The model takes in current images from surround-view cameras and the target slot to predict the locations of subsequent $n$ trajectory points in an autoregressive manner.
The controller steers the vehicle based on the path planning results, ego pose, and feedback signals to park the vehicle in the designated slot.
It is worth noting that the coordinates of the target point and predicted trajectory points are represented in the vehicle coordinate frame, ensuring the trajectory sequence and BEV features are expressed in consistent coordinate bases. This design also makes the entire system independent of the global coordinate frame.

Regarding the neural network details, the size of BEV features is $200\times200$, corresponding to the actual spatial range of $x\in[-10\mathrm{m}, 10\mathrm{m}]$, $y\in[-10\mathrm{m}, 10\mathrm{m}]$ with a resolution of $0.1$ meters.
In the transformer decoder, the maximum value of the trajectory serialization $N_t$ is $1200$.
The trajectory decoder generates a sequence of predictions with a length of $30$, achieving the best balance of accuracy and speed in inference.

We implement our method using the PyTorch framework.
The neural network is trained on one NVIDIA GeForce RTX 4090 GPU with a batch size of $16$, and the total training time is approximately $8$ hours with $40,000$ frames. Test data consists of about $5,000$ frames.

\subsection{Evaluation Metrics}
\subsubsection{\textbf{Model Trajectory Evaluation}}
To analyze the performance of a model before conducting a real scenario experiment, we design some evaluation metrics to evaluate the inference ability of the model.

\textbf{L2 Distance (L2 Dis.)}
L2 Distance refers to the average Euclidean distance between waypoints of the predicted and the ground-truth trajectories. This metric evaluates the precision and accuracy of model inference.

 \textbf{Hausdorff Distance (Haus. Dis.)} Hausdorff Distance refers to the maximum value of minimum distances between two point sets.
 This metric evaluates how well the predicted trajectory matches the ground-truth trajectory from the perspective of the point set.

 \textbf{Fourier Descriptor Difference (Four. Diff.)}
Fourier Descriptor Difference can be used to measure the difference between trajectories.
The lower value indicates the smaller difference between the trajectories.
This metric uses a certain number of Fourier descriptors to represent both the actual and predicted trajectories as vectors.
 
\subsubsection{\textbf{End-to-end Real-vehicle Evaluation}}
In real-vehicle experiments, we use the following metrics to evaluate end-to-end parking performance.

\textbf{Parking Success Rate (PSR)}
The parking success rate describes the probability of the ego vehicle successfully parking in the target parking slot.

\begin{figure*}[htbp]
	\centering
        \includegraphics[width=1\textwidth]{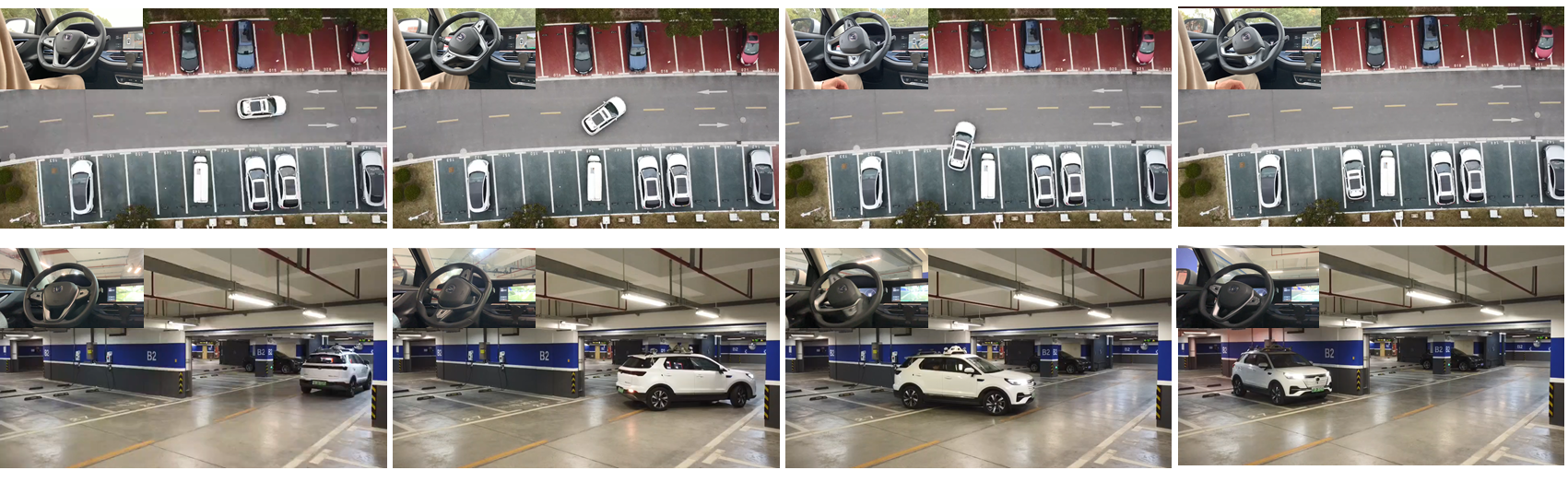}
	\caption{Illustration of the parking process across different scenarios. Each row showcases a parking case. Even in cases where there are obstacles such as cars or walls occupying adjacent parking spaces, our method can still effectively maneuver and park the vehicle in the designated spot.}
	\label{parking_show}
\vspace{-0.5cm}
\end{figure*}

\textbf{No Slot Rate (NSR)}
Rate of failure to park in designated parking spaces.

\textbf{Parking Violation Rate (PVR)}
Parking violation rate refers to a situation where a vehicle slightly extends beyond the designated parking space without obstructing or impeding adjacent parking spaces.

\textbf{Average Position Error (APE)}
The average position error is the average distance between the target parking position and the stopping position of the ego vehicle when parking successfully.

\textbf{Average Orientation Error (AOE)}
The average orientation error is the average difference between the target parking orientation and the stopping orientation of the ego vehicle when parking successfully.

\textbf{Average Parking Score (APS)}
The average parking score is calculated through a comprehensive evaluation that includes the position error, orientation error, and success rate during parking.
Scores are distributed between $0$ and $100$.

\textbf{Average Parking Time (APT)}
The average parking duration time across multiple parking maneuvers.
The parking duration is measured from the moment the parking mode is initiated until the vehicle is successfully parked in the designated space, or the parking process is terminated due to an anomaly or failure.


\subsection{Quantitative Results}
Using the proposed end-to-end parking system, we conducted closed-loop vehicle tests in four different parking garages to validate the performance of our proposed system. 
The results are shown in Table \ref{tab:vehicle_close_test}.

In the experiment, we tested in four different garages. Garage \uppercase\expandafter{\romannumeral 1} is an underground garage, and Garage \uppercase\expandafter{\romannumeral 2}, \uppercase\expandafter{\romannumeral 3} and \uppercase\expandafter{\romannumeral 4} are ground garages.
For each garage, we conducted three different experimental scenarios. Scene A is parking with no obstacles on either side. Scene B is parking with vehicles on the left side or right side. Scene C is parking with obstacles or walls nearby.
For each experimental scenario, we randomly selected three different parking slots. We conducted approximately three parking tests on both the left and right sides of each slot.
Experimental results show that our proposed method achieves a high parking success rate in different scenarios, exhibiting robust parking capability.

Despite the recent emergence of more end-to-end autonomous driving approaches, most of them concentrate on addressing challenges encountered in urban driving scenarios.
While methods such as ParkPredict  \cite{ParkPredict} are employed in parking scenarios, their tasks significantly differ from ours.
To the best of our knowledge, there is no existing effective end-to-end method that can be directly compared with our approach.
We compare the results of our method (the transformer-based decoder) and the Transfuser (the GRU-based decoder) in Table \ref{tab:exp_comparative}.
The transformer-based decoder has better prediction accuracy due to the attention mechanism in transformer.

\begin{table}[h]
\centering
\caption{Comparative Performance Evaluation}
\label{tab:exp_comparative}
\begin{tabularx}{\columnwidth}{@{}c|@{\extracolsep{\fill}}c@{\extracolsep{\fill}}c@{\extracolsep{\fill}}c@{}}
\toprule
Method & Haus. Dis. (m) $\downarrow$ & L2 Dis. (m) $\downarrow$ & Four. Diff. $\downarrow$ \\
\midrule
\textbf{Ours} & \textbf{0.076} & \textbf{0.033} & \textbf{0.43} \\
Transfuser \cite{chitta2022transfuser} & 0.676 & 0.458 & 11.51 \\
\bottomrule
\end{tabularx}
\vspace{0cm}
\end{table}

\subsection{Ablation Study}
We designed ablation experiments to analyze the impact of different network designs.
In terms of network structure, we conducted ablation experiments on the feature fusion, as illustrated in Table \ref{tab:ablation}.
We compare the results of the baseline (target query), feature concatenation, and feature element-wise addition.
The target query approach utilizes attention and spatial alignment mechanisms to fully integrate the target feature and the BEV feature.
It explicitly constrains the spatial relationship between the target slot and the BEV image to achieve the highest trajectory prediction accuracy.

\begin{table}[h]
\centering
\caption{Ablation Study on Feature Fusion}
\label{tab:ablation}
\begin{tabularx}{\columnwidth}{@{}c|@{\extracolsep{\fill}}c@{\extracolsep{\fill}}c@{\extracolsep{\fill}}c@{}}
\toprule
Method & Haus. Dis. (m) $\downarrow$ & L2 Dis. (m) $\downarrow$ & Four. Diff. $\downarrow$\\
\midrule
\textbf{Baseline} & \textbf{0.076} & \textbf{0.033} & \textbf{0.43} \\
Concatenation & 0.098 & 0.045 & 0.79 \\
Element-wise  & 0.097 & 0.047 & 0.83 \\
\bottomrule
\end{tabularx}
\vspace{0cm}
\end{table}

\subsection{Visualization}

The parking processes in different scenarios are shown in Fig. \ref{parking_show}, demonstrating the versatile adaptation capabilities across diverse scenarios of our algorithm. 



\subsection{Limitations}
Although our proposed method demonstrates advantages in the parking task, there are still some limitations.
First, our method has poor adaptability to moving targets due to the restriction of data scale and scenario diversity.
The model's adaptability to moving objects can be subsequently enhanced by expanding the dataset. Secondly, due to the training process that utilizes expert trajectories, it is impossible to provide effective negative samples.
Additionally, there is no robust corrective mechanism when a significant deviation occurs during the parking process, ultimately resulting in parking failure.
Subsequently, an end-to-end model can be trained using deep reinforcement learning by constructing a simulator that closely resembles real-world conditions through the utilization of NeRF \cite{mildenhall2021nerf} (Neural Radiance Field) and 3DGS \cite{kerbl20233d} (3D Gaussian Splatting).
Lastly, although our end-to-end parking method has achieved favorable results, there remains a gap compared to traditional rule-based parking methods.
However, we believe this problem will be solved as end-to-end technology continues to advance.
We expect that end-to-end parking algorithms will demonstrate advantages in complex scenarios in the future.

\section{Conclusion}
In this paper, we propose a camera-based end-to-end parking model.
The model inputs the target slot and the surround-view RGB images, obtains the fused features in BEV view by target query, and predicts the trajectory points using a transformer decoder in an autoregressive manner.
The results of trajectory planning are subsequently utilized for control.
We extensively evaluated the proposed method across various scenarios, and the results demonstrate its reliability and generalizability. 
Nevertheless, there still exists a performance gap between our end-to-end method and highly optimized rule-based parking methods.
In our future work, we aim to further improve the performance of end-to-end parking algorithms, with the expectation that learning-based approaches will eventually outperform traditional methods.
We believe that our research and practice will inspire and provoke thoughts among fellow researchers and engineers.

\bibliography{paper_tex}

\end{document}